\documentclass[journal,twoside,web]{ieeecolor}
\usepackage{tmi}
\usepackage{cite}
\usepackage{amsmath,amssymb,amsfonts}
\usepackage{algorithmic}
\usepackage{graphicx}
\usepackage{textcomp}

\makeatletter
\let\NAT@parse\undefined
\makeatother
\usepackage{hyperref}
\usepackage{multirow}
\usepackage{makecell}
\usepackage{booktabs}
\usepackage{array}
\usepackage{pifont}

\def\BibTeX{{\rm B\kern-.05em{\sc i\kern-.025em b}\kern-.08em
    T\kern-.1667em\lower.7ex\hbox{E}\kern-.125emX}}
\begin{document}
\title{EVOKE: Elevating Chest X-ray Report Generation via Multi-View Contrastive Learning and Patient-Specific Knowledge}
\author{Qiguang Miao \IEEEmembership{Senior Member, IEEE}, Kang Liu, Zhuoqi Ma, Yunan Li, Xiaolu Kang, Ruixuan Liu, Tianyi Liu, Kun Xie, and Zhicheng Jiao \IEEEmembership{Member, IEEE} 
\thanks{This work has been submitted to the IEEE for possible publication. Copyright may be transferred without notice, after which this version may no longer be accessible.}
\thanks{The work was jointly supported by the National Science and Technology Major Project under Grant No. 2022ZD0117103, the National Natural Science Foundations of China under Grant No. 62272364, the provincial Key Research and Development Program of Shaanxi under Grant No. 2024GH-ZDXM-47, the Research Project on Higher Education Teaching Reform of Shaanxi Province under Grant No. 23JG003, and the Fundamental Research Funds for the Central Universities under Grant No. ZYTS24090. \textit{(Corresponding authors: Kang Liu and Zhuoqi Ma)}}
\thanks{Qiguang Miao, Kang Liu, Zhuoqi Ma, Yunan Li, Xiaolu Kang, and Kun Xie are with the School of Computer Science and Technology, Xidian University, Xi'an, Shaanxi 710071, China (e-mail: \{qgmiao, zhuoqima, yunanli, xiekun\}@xidian.edu.cn; \{kangliu, 22031212472\}@stu.xidian.edu.cn).}
\thanks{Ruixuan Liu is with the Department of Orthopedics, Shanghai Key Laboratory for Prevention and Treatment of Bone and Joint Diseases, Shanghai Institute of Traumatology and Orthopedics, Ruijin Hospital, Shanghai Jiao Tong University School of Medicine, Shanghai 200025, China (e-mail: osteoliu@163.com).}
\thanks{Tianyi Liu is with the Department of Neurology, The First Affiliated Hospital of Harbin Medical University, Harbin, Heilongjiang 150001, China (e-mail: liutyi92@163.com).}
\thanks{Zhicheng Jiao is with the Department of Diagnostic Imaging, Brown University, Providence, RI 02903-4923, USA (e-mail: zhicheng\_jiao@brown.edu).}
}

\maketitle

\begin{abstract}
Radiology reports are crucial for planning treatment strategies and facilitating effective doctor-patient communication. However, the manual creation of these reports places a significant burden on radiologists. While automatic radiology report generation presents a promising solution, existing methods often rely on single-view radiographs, which constrain diagnostic accuracy. To address this challenge, we propose \textbf{EVOKE}, a novel chest X-ray report generation framework that incorporates multi-view contrastive learning and patient-specific knowledge. Specifically, we introduce a multi-view contrastive learning method that enhances visual representation by aligning multi-view radiographs with their corresponding report. After that, we present a knowledge-guided report generation module that integrates available patient-specific indications (e.g., symptom descriptions) to trigger the production of accurate and coherent radiology reports. To support research in multi-view report generation, we construct Multi-view CXR and Two-view CXR datasets using publicly available sources. Our proposed EVOKE surpasses recent state-of-the-art methods across multiple datasets, achieving a 2.9\% F\textsubscript{1} RadGraph improvement on MIMIC-CXR, a 7.3\% BLEU-1 improvement on MIMIC-ABN, a 3.1\% BLEU-4 improvement on Multi-view CXR, and an 8.2\% F\textsubscript{1,mic-14} CheXbert improvement on Two-view CXR.
\end{abstract}


\section{Introduction}
\label{sec:introduction}
\IEEEPARstart{R}{adiology} reports, crafted by experienced radiologists, meticulously document imaging findings from examinations such as X-rays, PET scans, and CTs, detailing abnormalities and initial diagnostic conclusions. These reports deliver vital imaging insights that empower physicians to develop efficient patient treatment strategies \cite{messina2022survey,tmi-2024-phraseaug}. However, the manual writing process is time-consuming and demands substantial expertise, posing challenges in meeting the increasing demands of modern healthcare \cite{bannur2024maira2groundedradiologyreport}, particularly in regions with limited medical resources. 

\begin{figure}
    \centering
    \includegraphics[width=1\linewidth]{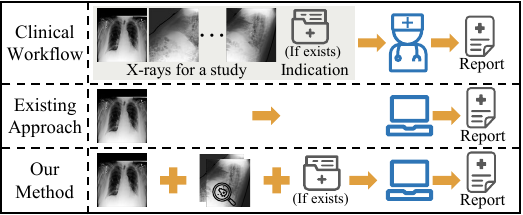}
    \caption{A comparison between existing methods and our proposed approach for chest X-ray report generation reveals that existing methods rely on single-view images, whereas our approach leverages multi-view radiographs and patient-specific indications.}
    \label{fig:0-introduction}
\end{figure}

Automatic chest X-ray report generation aims to produce detailed and accurate free-text reports based on multi-view radiographs, helping radiologists improve diagnostic efficiency and consistency by providing high-quality draft reports. Limitations like X-ray equipment constraints and the complexity of human anatomical structures can prevent a single-view radiograph from achieving optimal imaging quality and adequately displaying the overall anatomical structure. As a result, multi-view imaging examinations, such as postero-anterior (PA), antero-posterior (AP), and lateral views, are essential for accurate diagnostics and personalized treatment. In existing public datasets (e.g., MIMIC-CXR \cite{johnson-mimic-cxr-jpg} and MIMIC-ABN \cite{mimic-abn-ori}), each study includes a variable number of radiographs, an associated radiology report, and a patient-specific \textit{INDICATION} (which describe the patient's symptoms and may sometimes be missing). In the clinical workflow, radiologists typically select key X-rays as the primary diagnostic basis, supplementing them with additional X-rays and the patient's clinical symptoms (i.e., \textit{INDICATION}) to conduct a comprehensive analysis and generate a radiology report, as shown in Fig. \ref{fig:0-introduction}. However, existing methods \cite{wang_tmi_multicriteria_2022,liu_tmi_multi_grained_2024,li-dcl,aaai-liu2024bootstrapping-llm,liang2024divide-DCG} primarily generate reports based on single-view images, neglecting the detailed anatomical information available in multi-view images and the patient’s clinical symptoms. This limitation often results in inaccuracies and inconsistencies in the generated reports.

To address this challenge and emulate radiologists' clinical practices, we propose a new framework named \textbf{EVOKE}, which incorporates multi-view images and patient-specific knowledge (i.e., \textit{INDICATION}) into the chest X-ray report generation process. More concretely, we present a multi-view contrastive learning method that utilizes a multi-view fusion module to integrate a variable number of X-rays from the same study. This approach enhances visual representation by maximizing semantic correspondences both among multi-view images within a study and between these images and their corresponding report. Additionally, we introduce a knowledge-guided report generation module that effectively integrates available patient-specific \textit{INDICATION} and employs a transition bridge network to mitigate embedding space discrepancies caused by the presence or absence of \textit{INDICATION}. Together, these components facilitate the generation of accurate and coherent radiology reports. We evaluate our proposed EVOKE on the MIMIC-CXR \cite{johnson-mimic-cxr-jpg}, MIMIC-ABN \cite{mimic-abn-ori,recap}, and our curated Multi-view CXR and Two-view CXR datasets. Extensive experiments confirm that EVOKE outperforms recent state-of-the-art methods. Our key contributions are as follows:

\begin{itemize}
    \item We introduce the EVOKE framework for chest X-ray report generation, which effectively integrates multi-view images and patient-specific knowledge to enhance the accuracy of radiology reports.
    \item We propose a multi-view contrastive learning method that improves visual representations and flexibly handles a variable number of images per study.
    \item We present a knowledge-guided report generation module that incorporates patient-specific \textit{INDICATION} and mitigates embedding space discrepancies caused by the presence or absence of \textit{INDICATION}.
    \item We curate Multi-view CXR and Two-view CXR datasets from two public sources, ensuring that each study includes multiple radiographs. These datasets support research on multi-view report generation, particularly in scenarios with variable or two-view setups.
\end{itemize}

\begin{figure*}
    \centering
    \includegraphics[width=1\linewidth]{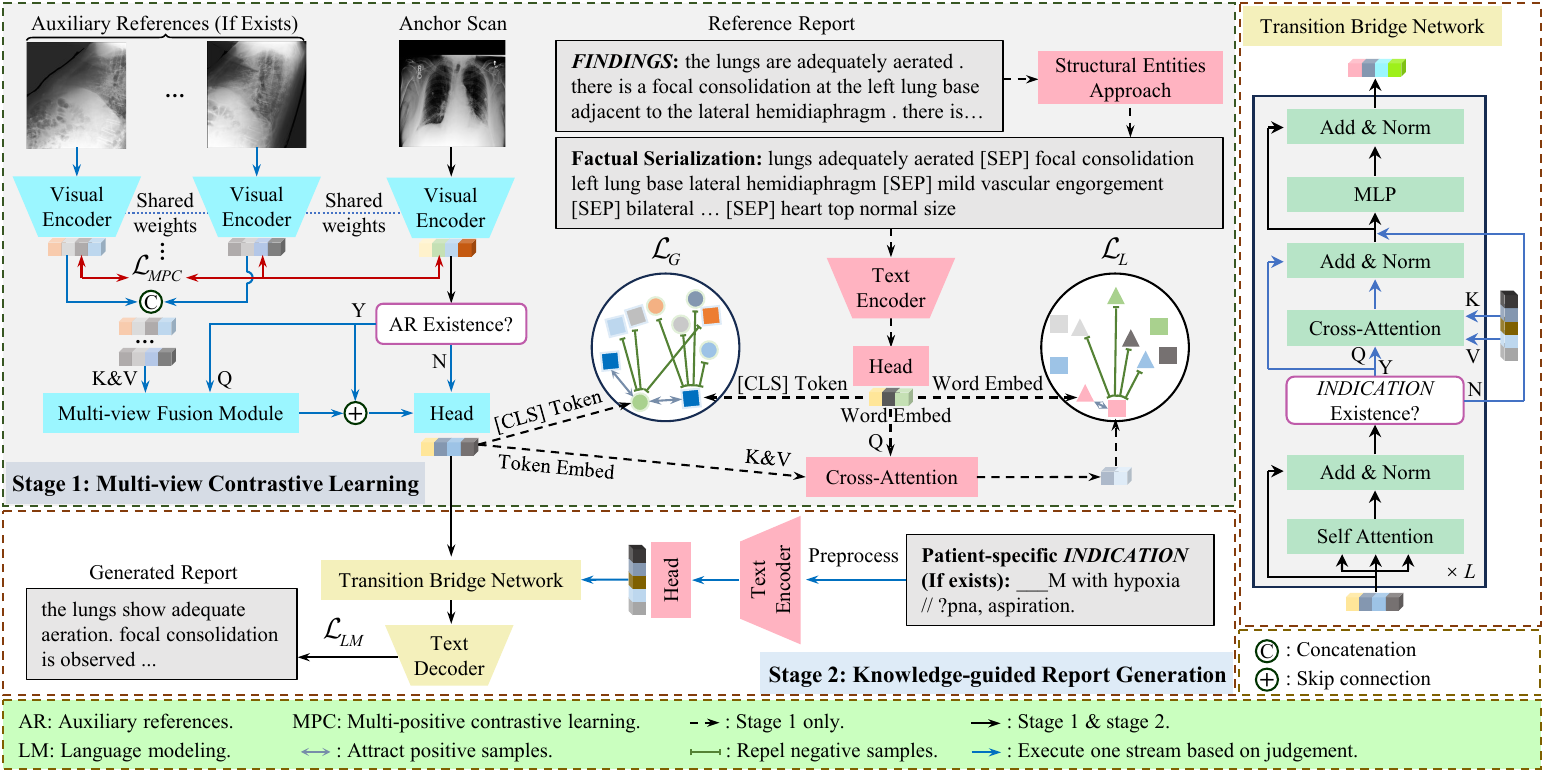}
    \caption{Illustration of our proposed EVOKE, which comprises a visual encoder, a text encoder, and a text generator. EVOKE employs a two-stage training strategy: Multi-view contrastive learning for representation learning (Stage 1) and knowledge-guided report generation (Stage 2). The model inference is performed solely using Stage 2.}
    \label{fig:overview}
\end{figure*}

\section{Related Work}
\subsection{Image Captioning}
Image captioning \cite{pmlr-blip,chen2023analog_bits} and radiology report generation \cite{2024-tmi-token-mixer,Bu_2024_CVPR,MLRG} both translate visual information into textual descriptions. The encoder-decoder framework has significantly advanced image captioning by organizing visual features into coherent text, benefiting radiology report generation. However, unlike image captioning, which typically generates concise and general descriptions, radiology report generation \cite{tmi-2024-organ} demands detailed, domain-specific, and diagnostically relevant content.


\subsection{Medical Image Analysis Meets Multi-view Learning}
Multi-view learning \cite{a_survey_multi_view,ZHAO2017-multi-view-survey} seeks to capture shared and complementary information from diverse perspectives of the same scene, thereby enhancing overall comprehension. In representation learning, \cite{paul_generalized_tmi} develop a multi-view semantic embedding extracted from X-ray reports, CT reports, and visual traits to enhance the clinical accuracy of chest X-ray diagnosis. REFERS \cite{Zhou_nature_2022} introduces a view fusion module that integrates fixed-view visual features, enhancing cross-modal alignment. In medical report generation, FMVP \cite{tmm_mulview_2024} incorporates single radiographs and auxiliary inputs (i.e., disease tags and medical concepts) as multi-view information, using two shared and continuous cross-attention mechanisms \cite{NIPS2017_transformer_attention} to assist in generating informative reports. Given that the IU X-ray dataset \cite{demner2016preparing} primarily comprises studies with two-view radiographs, several approaches \cite{chen-etal-2021-cross-modal,yang-m2kt} have been developed to handle two-view report generation. These methods enhance the clinical efficacy of generated reports and offer valuable insights for advancing multi-view report generation. However, they often struggle with the varying number of views encountered in real-world scenarios. To tackle this issue, we introduce a multi-view contrastive learning method that flexibly accommodates a variable number of images per study.

\subsection{Medical Image Analysis Meets Contrastive Learning}
Contrastive learning is pivotal in medical image analysis, particularly in tasks related to medical visual representation learning \cite{wang-mgca,cheng-prior} and report generation \cite{shen2024automatic_aaai,Deria_2024_CVPR}. It aligns radiographs with their corresponding reports by minimizing the distance between positive pairs (radiographs and associated reports) while maximizing the distance between negative pairs (radiographs and unrelated reports). Significant advancements have been made in this area. For instance, MedCLIP \cite{medclip_wang_2022} adopts a semantic matching loss for global alignment between decoupled radiographs and reports. ARL \cite{chen2022align} regards knowledge as an intermediary to facilitate semantic alignment between radiographs and their reports. MGCA \cite{wang-mgca} introduces multi-grained cross-modal alignment at the instance, pathological region, and disease levels for generalized medical visual representation. SEI \cite{sei} utilizes global and local cross-modal alignment between radiographs and factual serialization in reports to generate radiology reports. While these methods effectively align radiographs with their reports by extracting visual features from single-view radiographs, they often overlook the benefits of incorporating multi-view information from the same study. In response, our work introduces multi-view contrastive learning, integrating multi-view visual features to improve cross-modal alignment.

\section{Method}
\label{sec: method}
The overall architecture of our proposed EVOKE, shown in Fig. \ref{fig:overview}, adopts a two-stage training strategy. In Stage 1, we propose a multi-view contrastive learning method that integrates multi-view images to enhance visual representations. In Stage 2, we present a knowledge-guided report generation module that incorporates patient-specific \textit{INDICATION}, equipping the text generator with relevant background information to produce accurate and coherent radiology reports.

\subsection{Problem Formulation}
Let ${D_{tr}} =\left \{ ({{x_{i,k}},{x_{i,\backslash k}},{z_i},{r_i}}) \right \}_{i=1}^n$ be the training set, where \textit{n} denotes the number of studies and $k \in [1,m_i]$. The $i^{th}$ study includes \({m_i}\) radiographs (views), a patient-specific \textit{INDICATION} \(z_i\) (which may be absent), and a corresponding report \({r_i}\). To align with radiologists' clinical practices, we categorize the multi-view images within each study into an anchor scan \({x_{i,k}}\), serving as the primary diagnostic basis, and auxiliary references \({{x_{i,\backslash k}} = \left\{ {{x_{i,j}}\left| {j \ne k},{1 \le j \le {m_i}} \right.} \right\}}\), which provide additional context and may be empty. The \textit{INDICATION} \(z_i\) describes the patient's clinical symptoms; while not directly influencing diagnostic interpretation, it offers valuable contextual insights for radiograph analysis. Our objective is to learn a function \({F_\theta }\left(  \cdot  \right)\) that maps the input \(( {{x_{i,k}},{x_{i,\backslash k}},{z_i}} )\) to the corresponding report \({r_i}\) using the training set \({D_{tr}}\). 

\subsection{Multi-view Contrastive Learning Method}
\label{multi-view contrastive learning}
We introduce a multi-view contrastive learning method that enhances visual representations by utilizing semantic correspondences both among multi-view radiographs within a study and between these radiographs and their corresponding report. First, we employ multi-positive contrastive learning to align multi-view radiographs, improving visual feature consistency across them. Afterward, we develop a multi-view fusion module to integrate varying numbers of radiographs per study, generating fused visual features for subsequent cross-modal alignment. Finally, we apply contrastive learning at both instance-wise and token-wise levels, maximizing agreements between these radiographs and their corresponding report.

\textbf{Visual Features Extraction.} We use ResNet101 \cite{he-resnet}, pre-trained on ImageNet, as the visual encoder. The feature maps from the last convolutional layer of ResNet101 are regarded as the visual features of radiographs, formulated as \({\boldsymbol{V}} \in {\mathbb{R}^{M \times p \times d_1}}\), where \(M = \sum\nolimits_{i = 1}^B {{m_i}}\) denotes the total number of radiographs in the batch. Here, \(B\) is the batch size, \(p\) represents the dimensions of the feature map, and \(d_1\) corresponds to the number of channels. 

\textbf{Textual Features Extraction.} Inspired by \cite{yan2023style,fse}, we first employ the structural entities approach \cite{fse} to extract factual serialization from reports, mitigating noise for subsequent cross-modal alignment. Factual serialization, shown in Fig. \ref{fig:overview}, is a concise sentence constructed from clinically relevant keyword groups derived from reports. This serialization is then fed into a six-layer pre-trained SciBERT \cite{Beltagy2019} to produce textual features \({\boldsymbol{T}}\in {\mathbb{R}^{B \times k \times d_2}}\). Here, \(k\) denotes the number of textual tokens and \(d_2\) represents the dimensionality of each token. It is important to note that the textual features \({\boldsymbol{T}}\) are exclusively utilized in Stage 1.

\textbf{Multi-positive Contrastive Learning between Multi-view Radiographs}. To capture semantic correspondences between multi-view radiographs within the same study, we adopt multi-positive contrastive learning \cite{NEURIPS2023_stablerep}. This method aligns radiographs from the same study while differentiating them from those in other studies, thereby enhancing visual features consistency. We first excluded studies containing only a single-view radiograph from the batch (Notably, these studies are still used for subsequent cross-modal alignment). For  each anchor scan \({x_{i,a}}\), we compute the contrastive categorical distribution \(\boldsymbol{q}\) to quantify the similarity between \({x_{i,a}}\) and all other scans \({x_{\backslash a}}\). Here, \({x_{\backslash a}}\) comprises both auxiliary references of \({x_{i,a}}\) and radiographs from other studies. We denote the global visual features, after \(\ell_{2}\) normalization, as \(\boldsymbol{v}_{i,a}\) for anchor scans and \(\boldsymbol{v}_{\backslash a}\) for other scans. The $\boldsymbol{q} \in {\mathbb{R}^{K \times (K - 1)}}$ is defined as:
\begin{align}
{{\boldsymbol q}_i} = \frac{{\exp \left( {{{{{\boldsymbol v}_{i,a}} \cdot {{\boldsymbol v}_{\backslash a}^T}} \mathord{\left/ \right.
 } {{\tau _1}}}} \right)}}{{\sum\nolimits_{j = 1}^{K-1} {\exp \left( {{{{{\boldsymbol v}_{i,a}} \cdot {{({\boldsymbol{v}}_{\backslash a}^j)}^T}} \mathord{\left/ \right.
 } {{\tau _1}}}} \right)} }},
\end{align}
\noindent where \(K = \sum\nolimits_{j = 1,m_j>1}^B {{m_j}}\) denotes the total number of multi-view radiographs in a batch, and \({\tau _1} \in {{{\cal R}}^ + }\) is the temperature parameter. We then construct the ground-truth categorical distribution \({\boldsymbol{p}} \in {\mathbb{R}^{K \times (K - 1)}}\), formulated as:
\begin{align}
{\boldsymbol{p}_i} = \frac{{{\mathbb{I}_{match}}\left( {{x_{i,a}},{{{x}}_{\backslash a}}} \right)}}{{\sum\nolimits_{j = 1}^{K-1} {{\mathbb{I}_{match}}\left( {{{{x}}_{i,a}},{{x}}_{\backslash a}^j} \right)} }},
\end{align}

\noindent \({\mathbb{I}_{match}}\left( { \cdot , \cdot } \right)\) is an indicator function that determines whether two radiographs belong to the same study. Finally, the multi-positive contrastive loss is calculated using the cross-entropy between \(\boldsymbol{q}\) and \(\boldsymbol{p}\), denoted as: 
\begin{align}
{{{\cal L}}_{MPC}} =  - \frac{1}{K}\sum\limits_{i = 1}^{K} {{{\boldsymbol{p}}_i}\log {{\boldsymbol{q}}_i}}.
\end{align}
\noindent Despite the varying number of radiographs per study, the different number of non-zero elements in \({\boldsymbol{p}_i}\) account for this variability. Applying cross-entropy between \(\boldsymbol{q_i}\) and \(\boldsymbol{p_i}\) pushes the multi-view radiographs in the \(i^{th}\) study closer together in the embedding space.

\textbf{Multi-view Fusion Module.} Aligning multi-view radiographs with their corresponding reports necessitates effective integration of multi-view visual features. However, the varying number of images per study complicates simple concatenation methods, leading to inconsistent channel dimensions and hindering cross-modal alignment. While averaging and maximization schemes can ensure channel consistency, they sacrifice detailed information. To tackle this challenge, we propose a multi-view fusion module that flexibly integrates a variable number of multi-view images using the scaled dot-product cross-attention mechanism \cite{NIPS2017_transformer_attention}, denoted as \(ATTN\left ( Q,K\&V \right ) \). In this module, the anchor scan \({{x}_{i,a}}\) acts as queries, while auxiliary references \({{x}_{i,\backslash a}}\) serve as both keys and values. Skip connection \cite{he-resnet} and layer normalization (LN) are then applied to generate the fused visual features for subsequent cross-modal alignment, formulated as:
\begin{align}
{{\boldsymbol{\tilde{V}}}_{i,a}} = LN\left( {{{\boldsymbol{V}}_{i,a}} + ATTN\left( {{{\boldsymbol{V}}_{i,a}},{{\boldsymbol{V}}_{i,\backslash a}}} \right)} \right)
\end{align}
\noindent where \({\boldsymbol{\tilde V}} = \{ {{{{\boldsymbol{\tilde V}}}_{i,a}}\left| {1 \le i \le B} \right.} \} \in {\mathbb{R}^{B \times p \times d_1}}\) ensures consistent channel dimensions across studies while preserving information from both anchor scans and auxiliary references. Notably, we instruct the module to focus on processing one study at a time, flexibly accommodating the variability in the number of radiographs per study.

\textbf{Instance-wise Alignment Loss.} Drawing inspiration from \cite{cheng-prior,wang-mgca,fse}, we propose the instance-wise alignment loss to maximize agreements between multi-view radiographs and their corresponding report in the embedding space. We begin by projecting the fused visual features \({{\boldsymbol{\tilde{V}}}}\) and textual features \({\boldsymbol{T}}\) into a unified embedding space \(d\) using the two-layer convolutional projection head. The resulting embeddings are denoted as \({\boldsymbol{\bar V}} \in {\mathbb{R}^{B \times p \times d}}\) for visual features and \({\boldsymbol{\bar T}} \in {\mathbb{R}^{B \times k \times d}}\) for textual features. Subsequently, we define the visual-textual feature pair from the same study as a positive pair. To estimate the similarity between multi-view radiographs and their corresponding report, we compute the image-to-text contrastive categorical distribution \({{\boldsymbol{q}}^{v2t}} \in {\mathbb{R}^{B \times B}}\), given by:
\begin{align}
{{\boldsymbol{q}}^{v2t}} = \frac{{\exp \left( {{{{\boldsymbol{\bar v}} \cdot {{{\boldsymbol{\bar t}}}^T}} \mathord{\left/ \right.
 } {{\tau _2}}}} \right)}}{{\sum\nolimits_{j = 1}^B {\exp \left( {{{{\boldsymbol{\bar v}} \cdot {{({{{\boldsymbol{\bar t}}}^j})}^T}} \mathord{\left/ \right.
} {{\tau _2}}}} \right)} }},
\end{align}
\noindent where \({\boldsymbol{\bar v}}\) and \({\boldsymbol{\bar t}}\) are global visual and textual features processed by \(\ell_{2}\) normalization. Similarly, we compute the symmetric text-to-image contrastive categorical distribution \({{\boldsymbol{q}}^{t2v}}\). The global ground-truth categorical distribution \(\boldsymbol{p}^{g} \) is defined as:
\begin{align}
{\boldsymbol{p}}_{i,j}^{g} = \frac{{{\mathbb{I}_{equal}}\left( {r_i,r_j} \right)}}{{\sum\nolimits_{k = 1}^B {{\mathbb{I}_{equal}}\left( {{r_i},{r_k}} \right)} }},
\end{align}
\noindent where \(\boldsymbol{p}^{g} \in {\mathbb{R}^{B \times B}}\), and \({{\mathbb{I}_{equal}}\left( {\cdot,\cdot} \right)}\) denotes an indicator function that determines whether two reports are identical. The instance-wise alignment loss is expressed as:
\begin{align}
{{{\cal L}}_{G}} =  - \frac{1}{B}\sum\limits_{i = 1}^B {\left( {{\boldsymbol{p}}_i^{g}\log {\boldsymbol{q}}_i^{v2t} + {\boldsymbol{p}}_i^{g}\log {\boldsymbol{q}}_i^{t2v}} \right)}.
\end{align}
\noindent Our method differs from prior works \cite{cheng-prior,wang-mgca} in two key ways: We utilize factual serialization from reports for alignment instead of complete reports, and we employ multi-positive contrastive loss \cite{NEURIPS2023_stablerep} to learn semantic correspondences between multi-view radiographs and reports, rather than single-positive contrastive loss \cite{oord-cpc}.

\textbf{Token-wise Alignment Loss.} Inspired by \cite{wang-mgca,fse}, we employ the token-wise alignment loss \({{{\cal L}}_{L}}\) \cite{fse} to learn fine-grained visual features. This loss is achieved through single-positive contrastive learning between token embeddings of visual and textual data. In conclusion, the overall training objective in Stage 1 is formulated as:
\begin{align}
{{{\cal L}}_{pretrain}} = {{\cal L}}_{MPC} + {{\cal L}}_{G} + {{\cal L}}_{L}.
\end{align}

\begin{table*}
\centering
\begin{tabular}{c|ccc|ccc|ccc|ccc} 
\toprule
\multirow{2}{*}{\textbf{Split}} & \multicolumn{3}{c|}{\textbf{MIMIC-CXR}} & \multicolumn{3}{c|}{\textbf{MIMIC-ABN}} & \multicolumn{3}{c|}{\textbf{Multi-view CXR}} & \multicolumn{3}{c}{\textbf{Two-view CXR}} \\ 
\cline{2-13}
 & \textbf{\#Img} & \textbf{\#Rpt} & \textbf{\%Ind} & \textbf{\#Img} & \textbf{\#Rpt} & \textbf{\%Ind} & \textbf{\#Img} & \textbf{\#Rpt} & \textbf{\%Ind} & \textbf{\#Img} & \textbf{\#Rpt} & \textbf{\%Ind} \\ 
\midrule
Train & 269,239 & 150,957 & 66.4 & 69,526 & 34,763 & 64.6 & 220,978 & 100,505 & 67.1 & 165,056 & 82,528 & 67.9 \\
Val & 2,113 & 1,182 & 65.4 & 526 & 263 & 62.7 & 2,299 & 1,057 & 71.2 & 1,778 & 889 & 72.1 \\
Test & 3,852 & 2,343 & 57.3& 756 & 378 & 56.3 & 3,947 & 1,805 & 67.6 & 3,000 & 1,500 & 68.9 \\
\bottomrule
\end{tabular}
\caption{Statistics for the training, validation, and test sets across the four datasets. ``\#Img'' and ``\#Rpt'' denote the number of radiographs and reports, respectively. ``\%Ind'' represents the ratio of \textit{INDICATION}.}
\label{table: dataset}
\end{table*}

\begin{table*}
\centering
\begin{tabular}{c|c|c|c|cccccc|ccc} 
\toprule
\multirow{2}{*}{\textbf{Dataset}} & \multirow{2}{*}{\textbf{Method}} & \multirow{2}{*}{\textbf{Year}} & \multirow{2}{*}{\makecell{\textbf{Input} \\ \textbf{Size}}} & \multicolumn{6}{c|}{\textbf{NLG Metrics \(\uparrow\)}} & \multicolumn{3}{c}{\textbf{CE Metrics \(\uparrow\)}}  \\ 
\cmidrule{5-13}
 &  &  &  & \textbf{B-1} & \textbf{B-2} & \textbf{B-3} & \textbf{B-4} & \textbf{MTR} & \textbf{R-L} & \textbf{RG} & \textbf{CX5} & \textbf{CX14}\\ 
\midrule
\multicolumn{11}{l}{\textbf{Comparison with single-view methods}} \\ 
\midrule
\multirow{15}{*}{MIMIC-CXR} & R2Gen \cite{chen-etal-2020-generating}  & 2020 & 224 & 0.353 & 0.218 & 0.145 & 0.103 & 0.142 & 0.277 & 0.207 & 0.340 & 0.340 \\
 & CMN \cite{chen-etal-2021-cross-modal} & 2021 & 224 & 0.353 & 0.218 & 0.148 & 0.106 & 0.142 & 0.278 & 0.220 & 0.461 & 0.391 \\
 & CGPT2 \cite{nicolson-improving} & 2023 & 384 & 0.393 & 0.248 & 0.171 & 0.127 & 0.155 & 0.286 & - & - & 0.442 \\
 & MET \cite{wang2023metransformer} & 2023 & - & 0.386 & 0.250 & 0.169 & 0.124 & 0.152 & 0.291 & - & - & 0.311 \\
 & KiUT \cite{huang-kiut} & 2023 & 224 & 0.393 & 0.243 & 0.159 & 0.113 & 0.160 & 0.285 & - & - & 0.321 \\
 & SA \cite{yan2023style} & 2023 & 256  & - & 0.184 & - & - & - & - & 0.228 & - & 0.394 \\
 & FMVP \cite{tmm_mulview_2024} & 2023 & 224 & 0.389 & 0.236 & 0.156 & 0.108 & 0.150 & 0.284 & - & - & 0.336 \\
 & SEI-1 \cite{sei} & 2024 & 224 & - & 0.247 & - & 0.135 & 0.158 & 0.299 & 0.249 & 0.542 & 0.460 \\
 & MAN \cite{shen2024automatic_aaai} & 2024 & 224 & 0.396 & 0.244 & 0.162 & 0.115 & 0.151 & 0.274 & - & - & 0.389  \\
 & PMRG \cite{promtmrg-aaai-2024} & 2024 & 224  & \underline{0.398} & - & - & 0.112 & 0.157 & 0.268 & - & - & 0.476 \\
 & Med-LLM \cite{2024-mm-med-llm} & 2024 & 224  & - & - & - & 0.128 & 0.161 & 0.289 & - & - & 0.395 \\
 & HERGen \cite{2024-eccv-hergen} & 2024 & 384 & 0.395  & 0.248 & 0.169 & 0.122 & 0.156 & 0.285 & - & - & - \\ 
 & \textbf{EVOKE(Ours)} & - & 224  & 0.395 & \underline{0.262} & \underline{0.190} & \underline{0.147} & \underline{0.167} & \underline{0.311} & \underline{0.276} & \underline{0.557} & \underline{0.499} \\
 & \textbf{EVOKE(Ours)} & - & 384  & \textbf{0.408} & \textbf{0.271} & \textbf{0.197} & \textbf{0.151} & \textbf{0.171} & \textbf{0.313} & \textbf{0.278} & \textbf{0.578} & \textbf{0.517} \\
 & \(\Delta \uparrow\) & - & -  & +1.0\% & +2.1\% & +2.6\% & +1.6\% & +1.0\% & +1.4\% & +2.9\% & +3.6\% & +4.1\% \\ 
\midrule
\multirow{5}{*}{MIMIC-ABN} & R2Gen\(^\sharp\) \cite{chen-etal-2020-generating} & 2020 & 224 & 0.253 & 0.144 & 0.092 & 0.063 & 0.106 & 0.229 & 0.179 & 0.501 & 0.442  \\
 & CMN\(^\sharp\) \cite{chen-etal-2021-cross-modal} & 2021 & 224  & {0.256} & {0.147} & {0.095} & {0.066} & {0.110} & {0.230} & {0.183} & {0.528} & {0.460} \\
 & \textbf{EVOKE(Ours)} & - & 224  & \underline{0.310} & \underline{0.185} & \underline{0.125} & \underline{0.090} & \underline{0.127} & \underline{0.246} & \underline{0.214} & \underline{0.535} & \underline{0.482} \\
 & \textbf{EVOKE(Ours)} & - & 384  & \textbf{0.329} & \textbf{0.196} & \textbf{0.131} & \textbf{0.093} & \textbf{0.134} & \textbf{0.255} & \textbf{0.220} & \textbf{0.545} & \textbf{0.503} \\
 & \(\Delta \uparrow\) & - & - & +7.3\% & +4.9\% & +3.6\% & +2.7\% & +2.4\% & +2.5\% & +3.7\% & +1.7\% & +4.3\% \\ 
\midrule
\multirow{5}{*}{Multi-view CXR} & R2Gen\(^\sharp\) \cite{chen-etal-2020-generating} & 2020 & 224  & 0.359 & 0.225 & 0.155 & 0.114 & 0.143 & 0.297 & 0.255 & 0.431 & 0.384 \\
 & CMN\(^\sharp\) \cite{chen-etal-2021-cross-modal} & 2021 & 224  & {0.404} & {0.252} & {0.170} & {0.122} & {0.160} & {0.311} & {0.279} & {0.475} & {0.416}  \\
 & \textbf{EVOKE(Ours)} & - & 224  & \underline{0.413} & \underline{0.274} & \underline{0.199} & \underline{0.152} & \underline{0.174} & \underline{0.335} & \underline{0.328} & \underline{0.515} & \underline{0.487} \\
 & \textbf{EVOKE(Ours)} & - & 384  & \textbf{0.415} & \textbf{0.276} & \textbf{0.200} & \textbf{0.153} & \textbf{0.177} & \textbf{0.336} & \textbf{0.329} & \textbf{0.557} & \textbf{0.508} \\
 & \(\Delta \uparrow\) & - & -  & +1.1\% & +2.4\% & +3.0\% & +3.1\% & +1.7\% & +2.5\% & +5.0\% & +8.2\% & +9.2\%  \\ 
\midrule
\multicolumn{11}{l}{\textbf{Comparison with two-view methods}} \\ 
\midrule
\multirow{5}{*}{Two-view  CXR} & R2Gen\(^\sharp\) \cite{chen-etal-2020-generating} & 2020 & 224  & 0.346 & 0.219 & 0.153 & 0.113 & 0.141 & 0.302 & 0.267 & 0.413 & 0.400 \\
 & CMN\(^\sharp\) \cite{chen-etal-2021-cross-modal} & 2021 & 224  & {0.387} & {0.241} & {0.166} & {0.122} & {0.151} & {0.310} & {0.268} & {0.437} & {0.425}  \\
 & \textbf{EVOKE(Ours)} & - & 224  & \underline{0.393} & \underline{0.256} & \underline{0.184} & \underline{0.140} & \underline{0.165} & \underline{0.322} & \textbf{0.304} & \underline{0.531} & \underline{0.501} \\
 & \textbf{EVOKE(Ours)} & - & 384  & \textbf{0.411} & \textbf{0.270} & \textbf{0.195} & \textbf{0.150} & \textbf{0.172} & \textbf{0.326} & \underline{0.302} & \textbf{0.547} & \textbf{0.507} \\
 & \(\Delta \uparrow\) & - & -  & +2.4\% & +2.9\% & +2.9\% & +2.8\% & +2.1\% & +1.6\% & +3.6\% & +11.0\% & +8.2\%  \\

\bottomrule
\end{tabular}
\caption{Comparisons with SOTA methods on MIMIC-CXR, MIMIC-ABN, Multi-view CXR, and Two-view CXR datasets. \(\Delta\) denotes the performance difference between EVOKE and the best peer methods. \({\sharp}\) indicates results reproduced using official codes, while other results are cited from the original work. The best and second-best values are in \textbf{bold} and \underline{underlined}.}
\label{table: main-results}
\end{table*}

\subsection{Knowledge-guided Report Generation}
\textbf{Indication Features Extraction.} Patient-specific \textit{INDICATION}, as illustrated in Fig. \ref{fig:overview}, is collected before examinations and describes the patient's clinical symptoms. Although not directly tied to the diagnostic interpretation, it provides valuable contextual information that enhances the model's ability to analyze radiographs. However, de-identification in public datasets often introduces noise into \textit{INDICATION}, such as  \textit{\_\_\_}, \textit{@}, and \textit{-year-old}. We preprocess the data following the approach in \cite{sei}, which involves noise removal and standardization of gender expressions. The cleaned indications are then fed into the text encoder, initialized with the pre-trained model from Stage 1, to extract indication features.

\textbf{Report Generation.} As shown in Table \ref{table: dataset}, patient-specific \textit{INDICATION} (i.e., knowledge) may sometimes be absent in certain studies. Consequently, some existing methods either entirely ignore this data \cite{chen-etal-2020-generating,wang2023metransformer,shen2024automatic_aaai} or underutilize its potential \cite{tian-miccai-indication,ml4h-indication-rg,sei}. To address this limitation, we develop a transition bridge network, illustrated in Fig. \ref{fig:overview}. This network integrates the available \textit{INDICATION} via a cross-attention mechanism and utilizes the transition ``bridge'' to handle missing \textit{INDICATION}, thereby reducing embedding space inconsistencies. By uniformly processing cases with and without \textit{INDICATION}, the network supplies patient-specific knowledge to the text generator, facilitating the generation of more accurate and coherent reports. The text generator, implemented as a memory-driven transformer \cite{chen-etal-2020-generating}, produces radiology reports in an autoregressive manner, optimized by minimizing the negative log-likelihood as follows:
\begin{align}
{{{\cal L}}_{LM_i}} = - \sum\limits_{t = 1}^M {\log {{P}\left( {\left. {\tilde w_i^t} \right|{{x}_{i}},z_i,\tilde w^{j<t}_i} \right)}},
\end{align}
\noindent where \({\tilde w_i}\) denotes predicted words generated by the text generator. $x_i$ denotes the multi-view radiographs for the $i^{th}$ study, and \(z_i\) corresponds to the \textit{INDICATION}. Here, $M$ indicates the maximum number of generated tokens. 

\begin{table*}
\centering
\begin{tabular}{c|c|cccc|c|cccc|ccc} 
\toprule
\multirow{2}{*}{\textbf{Model}} & \multirow{2}{*}{\textbf{M/S}} & \multicolumn{4}{c|}{\textbf{Stage 1}} & \textbf{Stage 2} & \multicolumn{4}{c|}{\textbf{NLG Metrics $\uparrow$}} & \multicolumn{3}{c}{\textbf{CE Metrics $\uparrow$}} \\ 
\cmidrule{3-14}
 &  & \textbf{F/R} & \textbf{MPC} & \textbf{G} & \textbf{L} & \textbf{Ind (``bridge'')} & \textbf{B-2} & \textbf{B-4} & \textbf{MTR} & \textbf{R-L} & \textbf{RG} & \textbf{CX5} & \textbf{CX14} \\ 
\midrule
R2Gen & S & - & \ding{55} & \ding{55} & \ding{55} & \ding{55} (\ding{55}) & 0.218 & 0.103 & 0.142 & 0.277 & 0.207 & 0.340 & 0.340 \\
(a) & M & - & \ding{55} & \ding{55} & \ding{55} & \ding{51} (\ding{51}) & 0.247 & 0.134 & 0.158 & 0.296 & 0.252 & 0.545 & 0.480 \\
(b) & S & F & \ding{55} & \ding{51} & \ding{51} & \ding{55} (\ding{55}) & 0.213 & 0.102 & 0.142 & 0.278 & 0.233 & 0.512 & 0.430 \\
(c) & S & F & \ding{55} & \ding{51} & \ding{51} & \ding{51} (\ding{51}) & 0.242 & 0.130 & 0.158 & 0.297 & 0.256 & 0.548 & 0.484 \\
(d) & M & - & \ding{51} & \ding{55} & \ding{55} & \ding{51} (\ding{51}) & 0.261 & 0.146 & 0.165 & 0.309 & 0.270 & 0.551 & 0.482 \\
(e) & M & F & \ding{51} & \ding{51} & \ding{55} & \ding{51} (\ding{51}) & \textbf{0.263} & 0.145 & 0.166 & 0.305 & 0.268 & 0.540 & 0.480 \\
(f) & M & F & \ding{51} & \ding{55} & \ding{51} & \ding{51} (\ding{51}) & 0.246 & 0.136 & 0.157 & 0.305 & 0.255 & 0.513 & 0.450 \\
(g) & M & F & \ding{55} & \ding{51} & \ding{51} & \ding{51} (\ding{51}) & 0.252 & 0.142 & 0.162 & 0.307 & 0.266 & 0.536 & 0.477 \\
(h) & M & F & \ding{51} & \ding{51} & \ding{51} & \ding{55} (\ding{55}) & 0.226 & 0.108 & 0.148 & 0.279 & 0.230 & 0.539 & 0.477 \\
(i) & M & R & \ding{51} & \ding{51} & \ding{51} & \ding{51} (\ding{51}) & 0.258 & 0.142 & 0.164 & 0.307 & 0.266 & 0.522 & 0.470 \\ 
(j) & M & F & \ding{51} & \ding{51} & \ding{51} & \ding{51} (\ding{55}) & 0.259 & 0.145 & 0.164 & \textbf{0.312} & 0.272 & 0.548 & 0.488 \\ 
\midrule
\textbf{EVOKE} & M & F & \ding{51} & \ding{51} & \ding{51} & \ding{51} (\ding{51}) & 0.262 & \textbf{0.147} & \textbf{0.167} & 0.311 & \textbf{0.276} & \textbf{0.557} & \textbf{0.499} \\
\bottomrule
\end{tabular}
\caption{Ablation study on MIMIC-CXR with a resolution of $224^2$. ``M/S'' indicates a \textbf{m}ulti-view or \textbf{s}ingle-view method. ``F/R'' denotes cross-modal alignment using either \textbf{f}act serialization or \textbf{r}eport. ``MPC'', ``G'', and ``L'' indicate \textbf{m}ulti-\textbf{p}ositive \textbf{c}ontrastive loss, instance-wise alignment loss (\textbf{G}), and token-wise alignment loss (\textbf{L}), respectively. `Ind (``bridge'')' denotes whether \textit{INDICATION} is used in Stage 2 and if the transition ``bridge'' is added for handling missing \textit{INDICATION}.}
\label{table:ablation-study}
\end{table*}

\section{Experiments}
\subsection{Experimental Settings}
\textbf{Datasets.} We evaluate our approach on the following datasets: \textbf{1) MIMIC-CXR} \cite{johnson-mimic-cxr-jpg} is a large-scale chest X-ray dataset with free-text radiology reports and varying numbers of radiographs per study. \textbf{2) MIMIC-ABN} \cite{mimic-abn-ori}, a subset of MIMIC-CXR, focuses on abnormal sentences within radiology reports. \textbf{3) Our curated Multi-view CXR} aggregates studies with multiple views from both the MIMIC-CXR and IU X-ray \cite{demner2016preparing} datasets. \textbf{4) Our curated Two-view CXR} is a variant of Multi-view CXR that includes exactly two views per study, with other settings unchanged. Following established practices \cite{chen-etal-2020-generating,wang2023metransformer,shen2024automatic_aaai}, we treat the \textit{FINDINGS} section of each radiology report as the reference report, with MIMIC-ABN containing only abnormal sentences from this section. MIMIC-CXR and MIMIC-ABN datasets adhere to their official splits. For Multi-view CXR, we follow the MIMIC-CXR split for its MIMIC-CXR component and baseline split \cite{chen-etal-2020-generating,yang-m2kt,shen2024automatic_aaai} for its IU X-ray component. Detailed statistics for these datasets are summarized in Table \ref{table: dataset}.

\textbf{Evaluation Metrics.} We evaluate model performance using both natural language generation (NLG) and clinical efficacy (CE) metrics. NLG metrics measure the lexical similarities between generated and reference reports, including BLEU-\textit{n} (B-\textit{n}, where \(n \in \{ 1, 2, 3, 4 \} \)), METEOR (MTR), and ROUGE-L (R-L). CE metrics evaluate the clinical correctness of generated reports, including F\textsubscript{1} RadGraph (RG) \cite{jain-radgraph}, F\textsubscript{1,mic-5} CheXbert (CX5), and F\textsubscript{1,mic-14} CheXbert (CX14) \cite{Smit2020_chexbert}. RG metric quantifies the overlap of clinical entities and their relationships, aligning more closely with radiologists' evaluations than B-3 and CX14 metrics \cite{yu2023evaluating}. CX5 and CX14 metrics assess the model's ability to describe 5 and 14 clinical observations (e.g., \textit{Edema}, \textit{Pneumothorax}, and \textit{Cardiomegaly}) by computing the micro-averaged F\textsubscript{1} score for multi-label classification. These metrics are calculated using the pycocoevalcap \cite{chen2015microsoft-coco-caption}, radgraph \cite{jain-radgraph}, and f1chexbert \cite{Smit2020_chexbert} libraries, respectively.

\textbf{Implementation Details.} \textbf{1) MIMIC-CXR}: In stage 1, we train the model for 50 epochs using the AdamW optimizer with a learning rate of 5e-5. In stage 2, we use the RAdam optimizer for 50 epochs, with learning rates of 5e-6 for the pre-trained model and 5e-5 for other parameters. \textbf{2) MIMIC-ABN} and \textbf{Multi-view CXR}: As most images in these datasets are derived from MIMIC-CXR \cite{johnson-mimic-cxr-jpg}, Stage 1 is skipped, and the pre-trained model from MIMIC-CXR is fine-tuned using the RAdam optimizer for 50 epochs. Learning rates are set to 5e-6 for MIMIC-ABN \cite{mimic-abn-ori} and 5e-7 for Multi-view CXR. For fairness, all baselines are evaluated using models pre-trained on MIMIC-CXR and fine-tuned before assessment. \textbf{3) Two-view CXR}: The same optimizer, learning rate, and epoch settings as MIMIC-CXR are applied. Peer methods are implemented according to their official code, employing concatenation for visual feature fusion. \textbf{4) Common settings}: For EVOKE with a resolution of \(224^2\), radiograph pre-processing follows R2Gen \cite{chen-etal-2020-generating}. For \(384^2\) resolution, images are resized to \(448^2\), randomly cropped to \(384^2\), rotated by 5 degrees, and normalized. We exclude studies with empty or clinically insignificant reference reports (e.g., containing only ``\textit{Portable AP upright chest film \_\_\_ at 09:31 is submitted.}''). Stage 1 involves 242M parameters, while Stage 2 has 362M parameters. Experiments at $224^2$ resolution are conducted on an NVIDIA 3090 GPU (24GB), and those at $384^2$ resolution are performed on an NVIDIA A100 GPU (48GB). The batch size is set to 32. \(\tau_1\), \(\tau_2\), and the number of transition network blocks are set to 0.5, 0.5, and 1, respectively. The maximum number of generated tokens \(M\) is fixed at 100 across all datasets. The optimal model is selected based on the combined scores of B-4, RG, and CX14 on the validation set, with results reported on the test set.

\begin{table*}
\centering
\begin{tabular}{c|c|ccc|ccc|c|ccc|ccc} 
\toprule
\multirow{3}{*}{\textbf{Observation}}& \multicolumn{7}{c|}{\textbf{MIMIC-CXR}} & \multicolumn{7}{c}{\textbf{Two-view CXR}} \\ 
\cline{2-15}
 & \multirow{2}{*}{\textbf{\%}} & \multicolumn{3}{c|}{\textbf{R2Gen}} & \multicolumn{3}{c|}{\textbf{EVOKE (Ours)}} & \multirow{2}{*}{\textbf{\%}} & \multicolumn{3}{c|}{\textbf{R2Gen}} & \multicolumn{3}{c}{\textbf{EVOKE (Ours)}} \\ 
\cline{3-8}\cline{10-15}
 &  & \textbf{P} & \textbf{R} & \textbf{F\textsubscript{1}} & \textbf{P} & \textbf{R} & \textbf{F\textsubscript{1}} &  & \textbf{P} & \textbf{R} & \textbf{F\textsubscript{1}} & \textbf{P} & \textbf{R} & \textbf{F\textsubscript{1}} \\ 
\hline
ECM & 10.2 & 0.377 & \textbf{0.353} & 0.365 & \textbf{0.396} & 0.345 & \textbf{0.369} & 9.8 & \textbf{0.371} & 0.280 & 0.319 & 0.351 & \textbf{0.300} & \textbf{0.324} \\
Cardiomegaly & 14.7 & 0.567 & 0.347 & 0.431 & \textbf{0.628} & \textbf{0.566} & \textbf{0.595} & 14.1 & \textbf{0.590} & 0.409 & 0.483 & 0.578 & \textbf{0.567} & \textbf{0.573} \\
Lung Opacity & 13.3 & 0.457 & 0.231 & 0.307 & \textbf{0.578} & \textbf{0.352} & \textbf{0.437} & 13.1 & \textbf{0.614} & 0.069 & 0.124 & 0.567 & \textbf{0.335} & \textbf{0.421} \\
Lung Lesion & 2.6  & \textbf{0.481} & \textbf{0.046} & \textbf{0.084} & 0.217 & 0.036 & 0.061 & 2.8 & \textbf{0.500} & 0.012 & 0.023 & 0.357 & \textbf{0.060} & \textbf{0.102} \\
Edema & 8.3 & 0.445 & 0.254 & 0.323 & \textbf{0.514} & \textbf{0.447} & \textbf{0.478} & 5.8 & \textbf{0.523} & 0.259 & 0.346 & 0.475 & \textbf{0.333} & \textbf{0.392} \\
Consolidation & 3.3 & 0.188 & 0.096 & 0.127 & \textbf{0.240} & \textbf{0.154} & \textbf{0.188} & 3.5 & \textbf{0.300} & 0.115 & 0.167 & 0.279 & \textbf{0.163} & \textbf{0.206} \\
Pneumonia & 4.6 & 0.238 & \textbf{0.204} & \textbf{0.220} & \textbf{0.264} & 0.186 & 0.218 & 3.7 & 0.276 & 0.071 & 0.113 & \textbf{0.349} & \textbf{0.268} & \textbf{0.303} \\
Atelectasis & 11.1  & 0.421 & 0.249 & 0.313 & \textbf{0.496} & \textbf{0.519} & \textbf{0.507} & 10.2 & 0.463 & 0.248 & 0.323 & \textbf{0.483} & \textbf{0.464} & \textbf{0.473} \\
Pneumothorax & 1.0  & \textbf{0.333} & 0.029 & 0.054 & 0.291 & \textbf{0.157} & \textbf{0.204} & 0.8 & 0.000 & 0.000 & 0.000 & \textbf{0.222} & \textbf{0.087} & \textbf{0.125} \\
Pleural Effusion & 12.5 & \textbf{0.804} & 0.196 & 0.316 & 0.716 & \textbf{0.669} & \textbf{0.691} & 10.5 & \textbf{0.787} & 0.375 & 0.508 & 0.709 & \textbf{0.667} & \textbf{0.687} \\
Pleural Other & 1.7 & 0.000 & 0.000 & 0.000 & \textbf{0.217} & \textbf{0.083} & \textbf{0.120} & 1.9 & \textbf{0.500} & 0.018 & 0.034 & 0.091 & \textbf{0.036} & \textbf{0.051} \\
Fracture & 2.0 & 0.000 & 0.000 & 0.000 & \textbf{0.071} & \textbf{0.014} & \textbf{0.023} & 2.3 & 0.000 & 0.000 & 0.000 & \textbf{0.188} & \textbf{0.043} & \textbf{0.070} \\
Support Devices & 12.5 & 0.723 & 0.489 & 0.583 & \textbf{0.755} & \textbf{0.713} & \textbf{0.734} & 9.5 & 0.706 & 0.488 & 0.577 & \textbf{0.722} & \textbf{0.646} & \textbf{0.681} \\
No Finding & 2.2 & 0.130 & 0.626 & 0.215 & \textbf{0.223} & \textbf{0.660} & \textbf{0.333} & 12.0 & 0.387 & \textbf{0.950} & 0.550 & \textbf{0.547} & 0.813 & \textbf{0.654} \\ 
\hline
micro avg & - & 0.432 & 0.280 & 0.340 & \textbf{0.538} & \textbf{0.465} & \textbf{0.499} & - & 0.484 & 0.341 & 0.400 & \textbf{0.539} & \textbf{0.469} & \textbf{0.501} \\
macro avg & - & 0.369 & 0.223 & 0.238 & \textbf{0.400} & \textbf{0.350} & \textbf{0.354} & - & \textbf{0.430} & 0.235 & 0.255 & 0.423 & \textbf{0.342} & \textbf{0.362} \\
\bottomrule
\end{tabular}
\caption{Clinical accuracy of our EVOKE on the MIMIC-CXR and Two-view CXR datasets with a resolution of $224^2$. ECM stands for \textit{Enlarged Cardiomediastinum}. P, R, and F\textsubscript{1} denote the precision, recall, and F\textsubscript{1} score, respectively.}
\label{table:clinical_acc}
\end{table*}

\begin{figure*}
    \centering
    \includegraphics[width=1\linewidth]{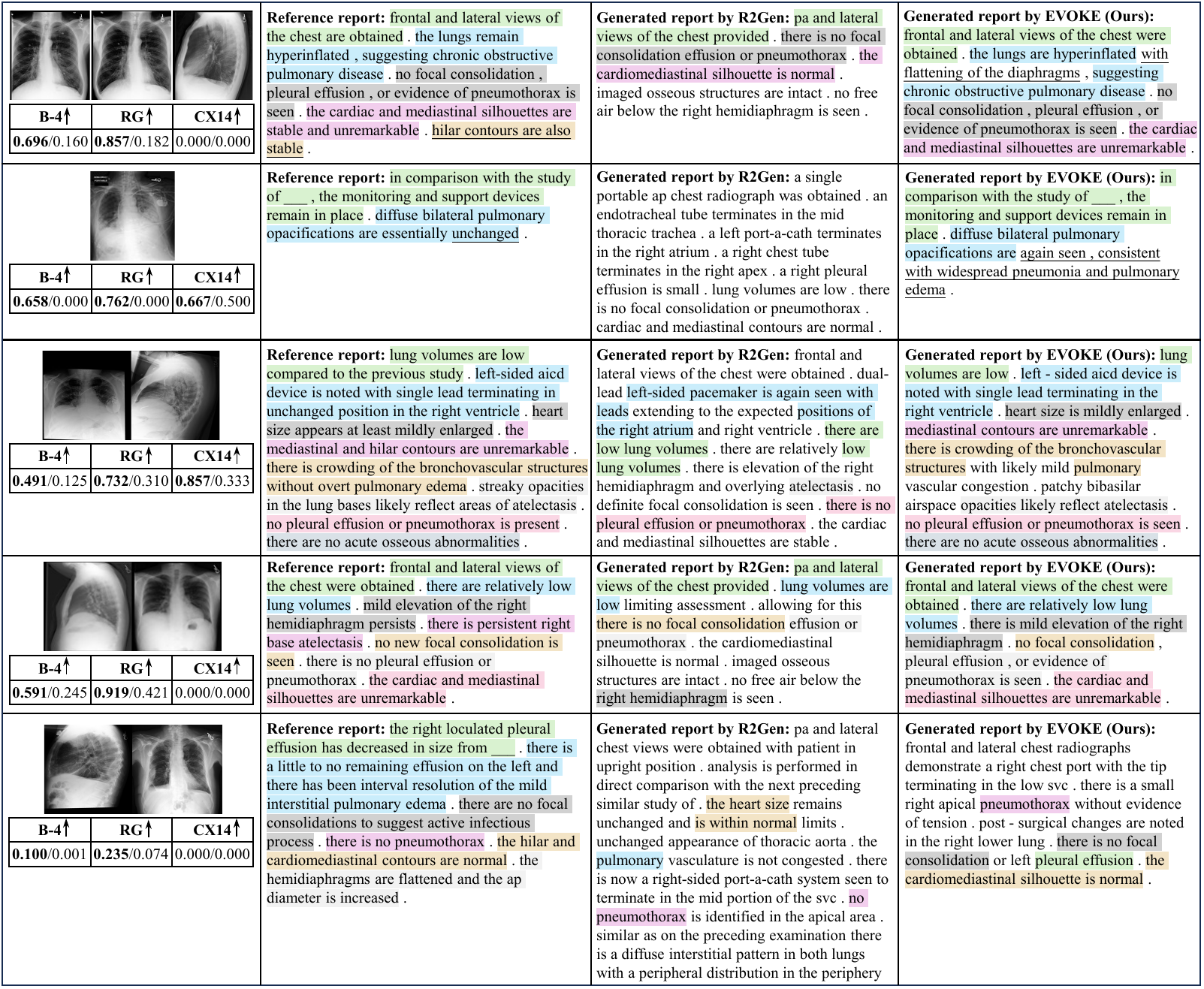}
    \caption{Examples of generated reports on the MIMIC-CXR test set with a resolution of $224^2$. The cell ``A/B'' represents ``EVOKE/R2Gen''. Sentences in the reference report are shown in different colors. Each sentence in generated reports is highlighted in matching colors corresponding to those in the reference report. Failure sentences in EVOKE are \underline{underlined}.}
    \label{fig:cases}
\end{figure*}

\begin{table}
\centering
\setlength{\tabcolsep}{1.8mm}
\begin{tabular}{c|cc|cc} 
\toprule
\textbf{Metric} & \textbf{w/ MV} & \textbf{w/o MV} & \textbf{w/ Ind} & \textbf{w/o Ind} \\ 
\midrule
\% (\textit{n}) & 70.7 (2,724) & 29.3 (1,128) & 57.8 (2,225) & 42.2 (1,627)  \\
B-1 & \textbf{0.406} & 0.369 & \textbf{0.413} & 0.372 \\
B-2 & \textbf{0.270} & 0.242 & \textbf{0.280} & 0.238 \\
B-3 & \textbf{0.196} & 0.176 & \textbf{0.206} & 0.169 \\
B-4 & \textbf{0.151} & 0.136 & \textbf{0.161} & 0.128 \\
MTR & \textbf{0.170} & 0.157 & \textbf{0.174} & 0.157 \\
R-L & \textbf{0.316} & 0.298 & \textbf{0.321} & 0.298 \\
RG & \textbf{0.287} & 0.249 & \textbf{0.299} & 0.244 \\
\bottomrule
\end{tabular}
\caption{Breakdown of EVOKE's metrics on the MIMIC-CXR test set using a resolution of $224^2$, categorized by: (I) the presence of multi-view radiographs (MV), and (II) the inclusion of an \textit{INDICATION} (Ind), in all studies. CX5-F and CX14-F denote the failure rates of the CX5 and CX14 metrics, respectively.}
\label{table:breakdown}
\end{table}

\subsection{Results and Analyses}
\textbf{Comparison with State-of-the-art Methods.} We compare our proposed EVOKE with 12 state-of-the-art (SOTA) methods: R2Gen \cite{chen-etal-2020-generating}, CMN \cite{chen-etal-2021-cross-modal}, CGPT2 \cite{nicolson-improving}, MET \cite{wang2023metransformer}, KiUT \cite{huang-kiut}, SA \cite{yan2023style}, FMVP \cite{tmm_mulview_2024}, SEI \cite{sei}, MAN \cite{shen2024automatic_aaai}, PMRG \cite{promtmrg-aaai-2024}, Med-LLM \cite{2024-mm-med-llm}, and HERGen \cite{2024-eccv-hergen}. Results are presented in Table \ref{table: main-results}, where ``\textit{Input Size}'' indicates the image resolution used by the visual encoder (e.g., 224 refers to a $224^2$ resolution). EVOKE outperforms all baselines across every metric on four datasets, as evidenced by a 2.9\%  F\textsubscript{1} RadGraph improvement on MIMIC-CXR, a 7.3\% BLEU-1 improvement on MIMIC-ABN, a 3.1\% BLEU-4 improvement on Multi-view CXR, and an 8.2\% F\textsubscript{1,mic-14} CheXbert improvement on Two-view CXR. These results highlight the effectiveness of EVOKE in generating clinically accurate reports. Additionally, at a resolution of \(384^2\), EVOKE consistently surpasses its performance at $224^2$ resolution on most metrics across these datasets, demonstrating that higher image resolution can significantly enhance the model's performance.

\textbf{Effect of Multi-view Contrastive Learning.} We evaluate the impact of our multi-view contrastive learning method on model performance, as shown in Table \ref{table:ablation-study}. Results show positive effects from multi-positive contrastive loss (EVOKE vs. (g)), instance-wise alignment loss (EVOKE vs. (f)), and token-wise alignment loss (EVOKE vs. (e)). EVOKE achieves superior performance across all metrics compared to (a) directly applying Stage 2 for report generation. This demonstrates the effectiveness of our multi-view contrastive learning method in enhancing visual features, thereby laying a solid foundation for Stage 2 to generate accurate and coherent reports based on patient-specific \textit{INDICATION}.

\textbf{Effect of Knowledge-guided Report Generation.} We assess the impact of our knowledge-guided report generation module on model performance, as detailed in Table \ref{table:ablation-study}. Both single-view ((c) vs. (b)) and multi-view methods (EVOKE vs. (h)) exhibit significant improvements across all metrics, emphasizing the importance of patient-specific \textit{INDICATION} in generating accurate reports. Additionally, EVOKE shows a 3.1\% increase in the sum of all metrics compared to (j), confirming the effectiveness of employing the transition ``bridge'' for handling missing \textit{INDICATION}.

\textbf{Clinical Accuracy of 14 Observations.} Table \ref{table:clinical_acc} illustrates the clinical accuracy of 14 observations related to thoracic diseases and support devices on the MIMIC-CXR and Two-view CXR datasets. Results reveal that: 1) EVOKE significantly improves most F\textsubscript{1} compared to R2Gen \cite{chen-etal-2020-generating}. 2) Although EVOKE is not specifically designed for imbalanced observations, it slightly outperforms the baseline, improving F\textsubscript{1} for \textit{Pleural Other} and \textit{Fracture} by 12\% and 2.3\% in MIMIC-CXR, and by 1.7\% and 7\% in Two-view CXR, respectively.

\textbf{Model Benefits from Multi-view Radiographs and \textit{INDICATION}.} Table \ref{table:breakdown} shows performance variations on subsets of the MIMIC-CXR test set with and without multi-view radiographs (\textit{INDICATION}). Significant improvements in NLG and RG metrics are observed for studies with multi-view radiographs (\textit{INDICATION}). This suggests that multi-view images offer rich visual information, while patient-specific knowledge provides essential contextual cues, collectively enhancing the accuracy of the generated radiology reports.


\textbf{Case Study.} Fig. \ref{fig:cases} presents five examples from the MIMIC-CXR test set to show the quality of our generated reports. A color distribution closely matching the reference reports indicates comprehensive coverage of clinical findings, while longer color bars in generated reports suggest more detailed descriptions. Results reveal that EVOKE generates higher-quality draft reports for radiologists compared to R2Gen \cite{chen-etal-2020-generating}. For instance, in case 1, radiologists only needed to add ``\textit{hilar contours are also stable}''. However, both EVOKE and R2Gen struggle to accurately capture disease progression due to the lack of consideration for patient temporal data, as seen in case 2, where the term ``\textit{unchanged}'' is not inferred.

\begin{figure}
    \centering
    \includegraphics[width=1\linewidth]{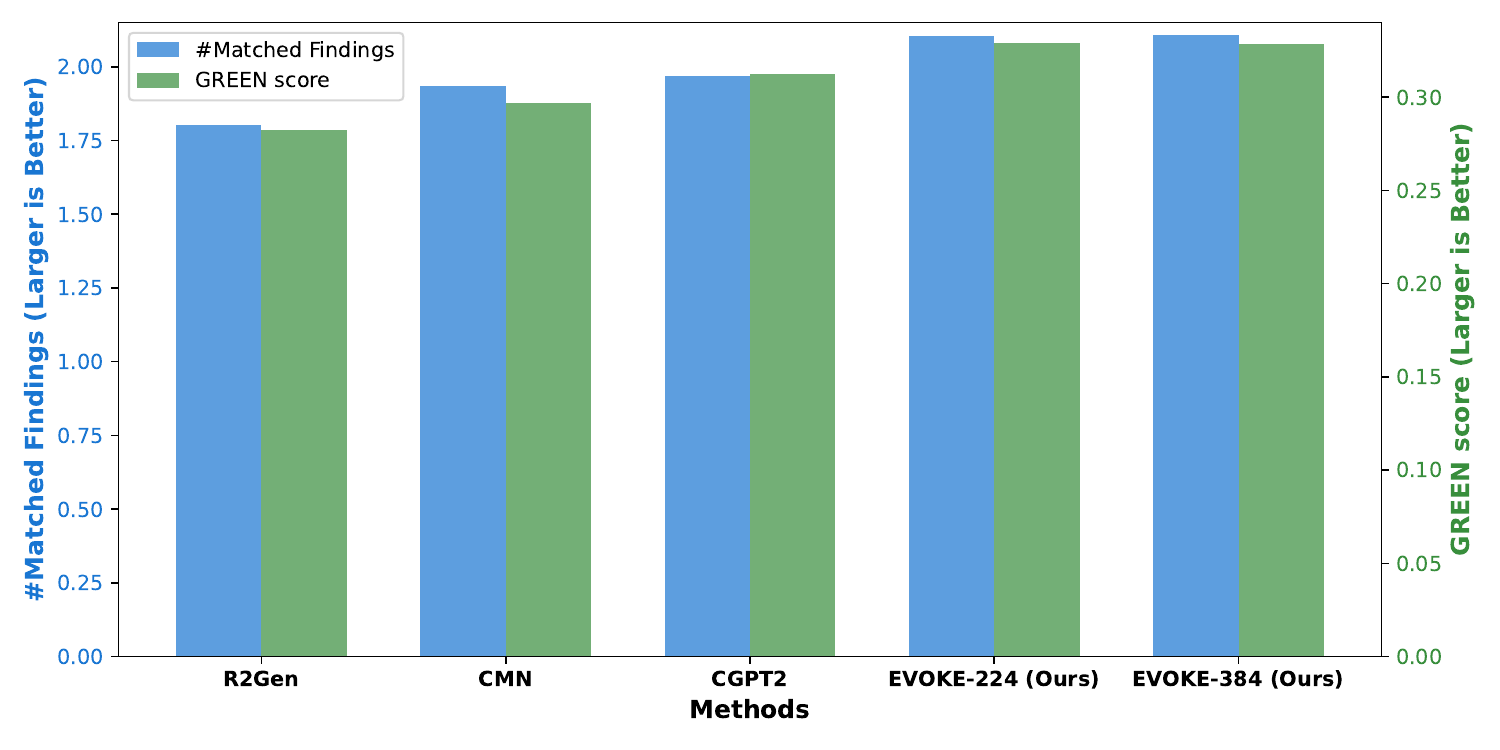}
    \caption{Comparisons with baselines on the MIMIC-CXR test set, evaluated by ``\#Matched Findings'' and GREEN score. ``EVOKE-s'' represents a resolution-specific variant of our proposed EVOKE framework.}
    \label{fig:llm-evaluate}
\end{figure}

\textbf{Evaluation via Large Language Models.} We evaluate the quality of our generated reports using the GREEN model \cite{2024-green}, which employs a fine-tuned LLaMA-2 \cite{llama-2-2023} model to detect clinically significant errors in radiology reports. The GREEN model identifies six types of clinically significant errors: (a) False report of a finding in the candidate; (b) Missing a finding present in the reference; (c) Misidentification of a finding's anatomic location/position; (d) Misassessment of the severity of a finding; (e) Mentioning a comparison that isn't in the reference; (f) Omitting a comparison detailing a change from a prior study. Based on the number of matched findings between generated and reference reports (denoted as ``\#Matched Findings''), the GREEN score is defined as:
\begin{align}
{\rm{GREEN}} = \frac{{{\rm{\#Matched}}\;{\rm{Findings}}}}{{{\rm{\#Matched }}\;{\rm{Findings}} + \sum\nolimits_{i = (a)}^{(f)} {{\rm{\#Error}}}_i }},
\end{align}
\noindent where ${{\#\rm{Error}}}_i$ refers to the count of the \(i^{th}\) clinically significant error. Fig. \ref{fig:llm-evaluate} compares our EVOKE with R2Gen \cite{chen-etal-2020-generating}, CMN \cite{chen-etal-2021-cross-modal}, and CGPT2 \cite{nicolson-improving} on the MIMIC-CXR test set, evaluated using ``\#Matched Findings'' and GREEN score. The result shows that EVOKE outperforms all baselines, indicating its superior capability to generate clinically accurate reports.

\begin{table*}
\centering
\setlength{\tabcolsep}{1.9mm}
\begin{tabular}{cccccccccccccccc} 
\toprule
\multirow{2}{*}{\textbf{ Model }} & \multicolumn{3}{c}{\textbf{(I)}} & \multicolumn{3}{c}{\textbf{(II)}} & \multicolumn{3}{c}{\textbf{(III)}} & \multicolumn{3}{c}{\textbf{(IV)}} & \multicolumn{3}{c}{\textbf{Overall Rank}} \\ 
\cmidrule(lr){2-4}\cmidrule(lr){5-7}\cmidrule(lr){8-10}\cmidrule(lr){11-13}\cmidrule(lr){14-16}
 & \textbf{ SE $\downarrow$} & \textbf{ IE $\downarrow$} & \textbf{ NE $\uparrow$} & \textbf{ SE $\downarrow$} & \textbf{ IE $\downarrow$} & \textbf{ NE $\uparrow$} & \textbf{ SE $\downarrow$} & \textbf{ IE $\downarrow$} & \textbf{ NE $\uparrow$} & \textbf{ SE $\downarrow$} & \textbf{ IE $\downarrow$} & \textbf{ NE $\uparrow$} & \textbf{ 1 } & \textbf{ 2 } & \textbf{ 3 } \\ 
\midrule
R2Gen & 24\% & 18\% & 58\% & 54\% & \textbf{38\%} & 8\% & 24\% & \textbf{2\%} & 74\% & 32\% & 32\% & 36\% & 22\% & 46\% & 32\% \\
CMN & 26\% & 24\% & 50\% & 34\% & 46\% & 20\% & 22\% & 12\% & 66\% & 42\% & \textbf{28\%} & 30\% & 26\% & 48\% & 26\% \\
EVOKE (Ours) & \textbf{16\%} & \textbf{14\%} & \textbf{70\%} & \textbf{24\%} & 44\% & \textbf{32\%} & \textbf{10\%} & 14\% & \textbf{76\%} & \textbf{12\%} & 34\% & \textbf{54\%} & \textbf{68\%} & 18\% & 14\% \\
\bottomrule
\end{tabular}
\caption{Human evaluation results, including severity assessment of four error types (SE: significant error, IE: insignificant error, NE: no error) and overall ranking of candidate reports (1: best, 3: worst).}
\label{table:human-evaluation}
\end{table*}

\textbf{Human Evaluations.} To ensure objectivity and reliability, we conduct a double-blind human evaluation of radiology report generation. We randomly select 50 cases from the MIMIC-CXR test set, each consisting of a reference report, multi-view radiographs, and three randomly ordered candidate reports generated by R2Gen \cite{chen-etal-2020-generating}, CMN \cite{chen-etal-2021-cross-modal}, and our EVOKE-224. Two experienced radiologists independently assess the candidate reports based on two key criteria: 1) Error type assessment. Each candidate report is evaluated for the severity of four potential errors: (I) False prediction of finding, (II) Omission of finding, (III) Incorrect location/position of finding, and (IV) Incorrect severity of finding. The severity is categorized into significant error (SE), insignificant error (IE), or no error (NE). 2) Candidate report overall ranking. The radiologists rank the three candidate reports in each case based on overall quality, assigning 1 (best), 2 (second-best), and 3 (worst). Table \ref{table:human-evaluation} presents the results, showing that our EVOKE consistently outperforms both R2Gen \cite{chen-etal-2020-generating} and CMN \cite{chen-etal-2021-cross-modal} across all error categories and overall ranking. Specifically, EVOKE achieves the lowest rates of significant errors (SE) and the highest rates of no errors (NE) for all error types (I-IV). Additionally, 68\% of EVOKE-generated reports are ranked as the best (1), significantly surpassing R2Gen (22\%) and CMN (26\%). These results indicate that EVOKE generates more accurate and clinically reliable radiology reports than baseline models.

\section{Conclusion}
In this paper, we proposed EVOKE for chest X-ray report generation by incorporating multi-view images and patient-specific \textit{INDICATION}. We introduced a multi-view contrastive learning method to align multi-view radiographs with their corresponding report, addressing the challenge of handling varying numbers of images per study. Additionally, we presented a knowledge-guided report generation module that integrates available patient-specific \textit{INDICATION}, providing the model with contextual insights. Experiments on MIMIC-CXR, MIMIC-ABN, Multi-view CXR, and Two-view CXR datasets show that incorporating multi-view radiographs, \textit{INDICATION}, and high-resolution radiographs improves the model's accuracy in describing clinical findings. In the future, we plan to utilize patient temporal data \cite{organ,2024-eccv-hergen} to capture disease progression and enhance model reliability through prediction uncertainty \cite{uncertainty}.

\bibliographystyle{IEEEtran}
\bibliography{ref}
\end{document}